\documentclass{article}

\usepackage[utf8]{inputenc}
\usepackage[T1]{fontenc}

\usepackage{amsmath}
\usepackage{amssymb}
\usepackage{amsfonts}

\usepackage[preprint]{neurips_2026}

\usepackage{amsthm}
\usepackage{booktabs}
\usepackage{algorithm}
\usepackage{algpseudocode}
\usepackage{microtype}
\usepackage{xcolor}
\usepackage{tikz}
\usetikzlibrary{arrows.meta, positioning}
\usepackage{url}
\usepackage{hyperref}

\begin{document}


\title{RotaryQuant: Fitting 120B MoE Models on Consumer Hardware via
Fused Compressed-Space Attention}

\author{%
  A.~Lui\textsuperscript{1} \qquad M.~Elsaied\textsuperscript{1,3} \qquad
  N.~P.~Savani\textsuperscript{2,4} \\[5pt]
  \normalfont\small
  \begin{minipage}{0.95\textwidth}\centering
    \textsuperscript{1}Cognizant AI \& Analytics, London, UB8 3PH, UK\\
    \textsuperscript{2}Cognizant AI Lab, Responsible AI Office, San Francisco, USA\\
    \textsuperscript{3}Southern Methodist University (SMU), Engineering Management,
    Lyle School of Engineering, Dallas, TX, USA\\
    \textsuperscript{4}University of Maryland, Baltimore County, Goddard Planetary
    Heliophysics Institute, 5523 Research Park Drive, Baltimore, MD 21228, USA
  \end{minipage}%
}

\maketitle

\begin{abstract}
Large mixture-of-experts (MoE) language models with 26--120 billion parameters
exceed the memory capacity of consumer devices through three simultaneous
pressures: resident weight matrices, key-value (KV) cache state that grows
linearly with context, and dozens of expert sublayers that must be paged on
demand. We present RotaryQuant, a three-axis compression system that addresses
all three. Mixed-precision weight quantization assigns bit-widths by
architectural role: 4-bit for dense layers, 2-bit for routed experts, and 8-bit
for the shared expert whose high activation kurtosis resists aggressive
compression. LRU expert offloading pages non-resident experts to disk under
genuine memory pressure. The novel axis is IsoQuant, a KV cache compression
method that applies a Walsh--Hadamard transform followed by block-diagonal
SO(4) rotations to isotropize activation distributions before 3-bit scalar
quantization, requiring $O(d \log d)$ operations and 256 stored parameters per
head versus $O(d^2)$ and 16{,}384 for dense rotation methods. A fused four-kernel
Metal GPU pipeline performs attention directly on packed 3-bit tensors without
materializing full-precision KV state---a different execution model, not just a
quantization scheme. The combined system fits Gemma 4-26B-A4B and Qwen3-30B-A3B
within a 16\,GB budget and Nemotron-H 120B within 32\,GB, running interactively
at 9--19 tok/s with near-zero perplexity degradation ($\Delta$PPL $\leq +0.0012$)
and 100\% retrieval accuracy at 32K context.
\end{abstract}

\section{Introduction}

Mixture-of-Experts (MoE) architectures~\citep{shazeer2017outrageously,fedus2022switch}
have enabled language models to scale to hundreds of billions of parameters while
maintaining tractable inference costs through sparse activation. However, deploying
these models on consumer hardware---devices with 16--32\,GB of unified
memory---remains infeasible due to two simultaneous memory pressures. A
30B-parameter MoE at 4-bit precision requires ${\sim}15$\,GB for weights alone; an
8K-token context window adds another 4--8\,GB of key-value (KV) cache
state~\citep{vaswani2017attention}.

The consequences of cloud-only inference extend beyond cost. Each prompt leaves the
user's device, creating privacy exposure. Recurring API subscriptions exclude
lower-income users and disadvantage those in low-bandwidth regions. Datacenter
energy and water demand continue to grow faster than per-chip efficiency
gains~\citep{iea2024electricity,uptime2023water}. These factors---privacy,
accessibility, and environmental cost---motivate local inference as a first-class
deployment target.

\begin{figure}[t]
  \centering
  \begin{tikzpicture}[
    font=\scriptsize,
    box/.style={draw, rounded corners=1pt, minimum height=6.5mm, align=center,
      inner sep=2pt, text width=13mm},
    gbox/.style={box, draw=green!55!black, fill=green!8},
    arr/.style={-{Latex[length=1.3mm]}, semithick},
    xs/.style={node distance=0pt},
  ]
    \def\dx{1.92}
    \node[anchor=east] at (-0.15,0) {Standard:};
    \node[box] (s1) at (0.8,0) {Packed KV\\(3-bit)};
    \node[box] (s2) at (0.8+\dx,0) {Dequant\\$R^{-1}$};
    \node[box] (s3) at (0.8+2*\dx,0) {FP16\\tensors};
    \node[box] (s4) at (0.8+3*\dx,0) {GEMM\\$Q \cdot K^\top$};
    \node[box] (s5) at (0.8+4*\dx,0) {Softmax\\$+ \cdot V$};
    \node[box] (s6) at (0.8+5*\dx,0) {Output};
    \foreach \a/\b in {s1/s2, s2/s3, s3/s4, s4/s5, s5/s6} { \draw[arr] (\a) -- (\b); }
    \node[above=0.5mm of s3, text=black!70] {\itshape materialized};

    \node[anchor=east] at (-0.15,-1.35) {IsoQuant:};
    \node[gbox] (i1) at (0.8,-1.35) {Packed KV\\(3-bit)};
    \node[gbox] (i2) at (0.8+\dx,-1.35) {Fused\\QK dot};
    \node[gbox] (i3) at (0.8+2*\dx,-1.35) {Softmax};
    \node[gbox] (i4) at (0.8+3*\dx,-1.35) {Fused\\V accum};
    \node[gbox] (i5) at (0.8+4*\dx,-1.35) {$R^{-1}$\\($1\times$)};
    \node[gbox] (i6) at (0.8+5*\dx,-1.35) {Output};
    \foreach \a/\b in {i1/i2, i2/i3, i3/i4, i4/i5, i5/i6} { \draw[arr, green!55!black] (\a) -- (\b); }
    \node[below=0.5mm of i3, text=green!55!black] {\itshape no materialization};
  \end{tikzpicture}
  \caption{Standard decode reconstructs full FP16 tensors before GEMM-based attention.
  IsoQuant computes attention directly on packed 3-bit data via fused kernels, applying
  the inverse rotation only once to the aggregated output.}
  \label{fig:pipeline}
\end{figure}
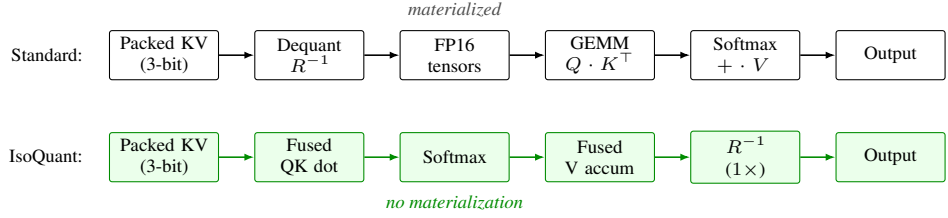

Prior work on KV cache compression~\citep{liu2024kivi,hooper2024kvquant,zandieh2025qjl,zandieh2026turboquant}
reduces cache memory through quantization, but universally follows a
reconstruct-then-compute paradigm: compressed KV entries are decompressed to full
precision before standard GEMM-based attention (Figure~\ref{fig:pipeline}, top).
This materialization step negates much of the memory savings during computation and
introduces bandwidth overhead proportional to sequence length.

We propose a fundamentally different execution model: compute-in-compressed-space
(Figure~\ref{fig:pipeline}, bottom). Rather than materializing FP16 tensors, our
fused Metal kernels perform QK dot products and value accumulation directly on 3-bit
packed data. The key enabler is IsoQuant, a structured rotation that spreads KV
vector energy isotropically across dimensions before quantization. By composing a
Walsh--Hadamard transform (WHT) with block-diagonal SO(4) rotations, IsoQuant
achieves $O(d \log d)$ rotation cost with only 256 stored parameters per head,
compared to $O(d^2)$ cost and 16{,}384 parameters for dense rotation
methods~\citep{zandieh2026turboquant,liu2024spinquant}.

Our contributions are:
\begin{enumerate}
  \item \textbf{IsoQuant rotation.} A WHT + SO(4) structured rotation that achieves
  isotropic energy spread with $O(d \log d)$ complexity and $64\times$ fewer stored
  parameters than dense alternatives (\S\ref{sec:isoquant}).
  \item \textbf{Fused compressed-space decode.} A four-kernel Metal pipeline that
  computes attention directly on packed 3-bit KV data, eliminating tensor
  materialization during autoregressive decode (\S\ref{sec:fused}).
  \item \textbf{RotaryQuant: a three-axis memory system.} A principled composition
  of weight quantization, KV cache compression (IsoQuant), and LRU expert offloading
  that enables 120B-parameter MoE inference within 17.2\,GB peak on a 32\,GB Apple M4
  Max (\S\ref{sec:threeaxis}).
  \item \textbf{Comprehensive evaluation.} Near-zero quality degradation
  ($\Delta$PPL $< 0.002$) across three architecturally distinct models, with decode
  profiling that quantifies when KV compression is and is not beneficial
  (\S\ref{sec:experiments}).
\end{enumerate}

\section{Related Work}

\paragraph{KV cache quantization.}
KIVI~\citep{liu2024kivi} applies asymmetric 2-bit quantization to keys and values,
demonstrating that keys benefit from per-channel quantization while values prefer
per-token schemes. KVQuant~\citep{hooper2024kvquant} extends this with non-uniform
quantization and dense-and-sparse decomposition to support 10M+ context windows.
QJL~\citep{zandieh2025qjl} projects KV vectors via Johnson--Lindenstrauss transforms
before 1-bit quantization. All methods reconstruct full-precision tensors before
attention computation.

\paragraph{Rotation-based compression.}
SmoothQuant~\citep{xiao2023smoothquant} smooths activation outliers via channel-wise
scaling to enable 8-bit quantization. QuaRot~\citep{ashkboos2024quarot} applies
Hadamard rotations to remove outliers before 4-bit weight and activation
quantization. SpinQuant~\citep{liu2024spinquant} learns dense rotation matrices
end-to-end. TurboQuant~\citep{zandieh2026turboquant} combines random dense rotations
with online vector quantization for KV caches, achieving strong compression but
requiring $d^2$ stored rotation parameters and $O(d^2)$ FMAs per application.
IsoQuant replaces the dense rotation with a structured WHT + SO(4) decomposition,
reducing both storage and compute by over an order of magnitude while enabling fused
decode. Table~\ref{tab:rotation} summarizes the key differences.

\begin{table}[t]
  \caption{Comparison of rotation-based KV compression methods at head dimension
  $d = 128$.}
  \label{tab:rotation}
  \centering
  \small
  \setlength{\tabcolsep}{4.5pt}
  \begin{tabular}{lcccc}
    \toprule
    Method & Rotation Type & KV Materialized & Decode FMAs & Stored Params \\
    \midrule
    TurboQuant~\citep{zandieh2026turboquant} & Dense random & Yes & $O(d^2)$ & 16{,}384 \\
    SpinQuant~\citep{liu2024spinquant} & Learned dense & Yes & $O(d^2)$ & 16{,}384 \\
    QuaRot~\citep{ashkboos2024quarot} & Hadamard & Yes & $O(d \log d)$ & 0 \\
    IsoQuant (ours) & WHT + SO(4) & No & $O(d \log d)$ & 256 \\
    \bottomrule
  \end{tabular}
\end{table}

\paragraph{Expert offloading.}
FlexGen~\citep{sheng2023flexgen} pioneered offloading strategies for single-GPU
inference, trading throughput for memory via CPU/disk swapping. We adapt this to
MoE-specific expert offloading with LRU eviction, composing it orthogonally with KV
compression.

\paragraph{Efficient attention.}
FlashAttention~\citep{dao2022flashattention} reduces attention memory from $O(n^2)$
to $O(n)$ through tiling, but operates on uncompressed KV. Our fused kernels can be
viewed as a compressed-domain analogue that trades exact GEMM for direct packed-data
arithmetic.

\section{Method}

\subsection{RotaryQuant: Three Compression Axes}
\label{sec:threeaxis}

Consumer-scale MoE inference must simultaneously address three memory consumers:
weight matrices $W$, KV cache state $(k_j, v_j)_{j=1}^{T}$, and expert residency.
RotaryQuant composes three independent compression mechanisms in canonical order:
\begin{equation}
  M_{\text{total}} = O_E \circ C_{KV} \circ Q_W
\end{equation}
where $Q_W$ is weight quantization, $C_{KV}$ is KV cache compression, and $O_E$ is
expert offloading. This composition is non-commutative: weight quantization
determines the activation distribution that KV compression must handle, and expert
offloading depends on the post-compression memory budget.

\paragraph{Weight quantization.}
Dense layers use 4-bit (GPTQ/AWQ~\citep{frantar2023gptq,lin2024awq}); routed MoE
experts use 2-bit; the shared expert uses Q8\_0 due to high activation kurtosis
($\kappa = 10.10$ vs.\ $0.41$ for specialist experts).

\paragraph{Expert offloading.}
Non-resident experts are evicted to disk via LRU policy and loaded on demand,
trading latency for memory. This is beneficial only under genuine memory pressure
(see Section~\ref{sec:offload}).

\subsection{IsoQuant Compression Pipeline}
\label{sec:isoquant}

Each key or value vector $x \in \mathbb{R}^d$ ($d=128$ for standard attention heads)
undergoes four stages, summarized in Algorithm~\ref{alg:isoquant}:

\begin{algorithm}[t]
  \caption{IsoQuant KV Compression}
  \label{alg:isoquant}
  \begin{algorithmic}[1]
    \Require KV vector $x \in \mathbb{R}^d$, WHT matrix $H_d$, SO(4) quaternion pairs
    $\{(q_L^{(i)}, q_R^{(i)})\}_{i=1}^{d/4}$, codebook $\mathcal{C}_b$
    \Ensure Packed representation $\bar{x}$, scale $s$
    \State $s \gets \lVert x \rVert_2$;\quad $\hat{x} \gets x/s$ \Comment{Normalize to unit sphere}
    \State $\tilde{x} \gets H_d \cdot \hat{x}$ \Comment{WHT global mix: $O(d \log d)$}
    \For{$i = 1$ to $d/4$}
      \State $\tilde{x}_{4i:4i+4} \gets q_L^{(i)} \cdot \tilde{x}_{4i:4i+4} \cdot \bar{q}_R^{(i)}$ \Comment{SO(4) block rotation}
    \EndFor
    \State $\bar{x} \gets \textsc{BitPack}_b(\textsc{LloydMax}(\tilde{x}, \mathcal{C}_b))$ \Comment{Quantize + pack to $b$ bits}
    \State \Return $\bar{x}, s$
  \end{algorithmic}
\end{algorithm}

\paragraph{Step 1: Normalize.}
Map to the unit sphere: $\hat{x} = x/\lVert x \rVert_2$, storing $\lVert x \rVert_2$
as a scalar scale factor.

\paragraph{Step 2: Structured rotation.}
Apply the composed rotation $R = R_{SO(4)} \circ H_d$:
\begin{equation}
  \tilde{x} = R_{SO(4)}(H_d \cdot \hat{x})
\end{equation}
where $H_d$ is the $d \times d$ normalized Walsh--Hadamard matrix and $R_{SO(4)}$
applies independent SO(4) rotations to each contiguous 4-dimensional block.

The WHT provides global mixing at cost $O(d \log d)$ via its butterfly structure,
spreading energy that may be concentrated in a few dimensions. The subsequent SO(4)
block rotations perform fine-grained alignment within each 4-element group,
parameterized by paired unit quaternions $(q_L, q_R)$.

\paragraph{Step 3: Scalar quantization.}
Each coordinate of $\tilde{x}$ is quantized to $b$ bits using Lloyd--Max optimal
codebooks~\citep{lloyd1982least} that minimize expected distortion under the
post-rotation marginal distribution.

\paragraph{Step 4: Bit-pack.}
At $b = 3$ bits, $d = 128$ dimensions pack into 48 bytes, achieving ${\sim}5.3\times$
compression versus 256 bytes in FP16.

\paragraph{Why structured rotation works.}
The WHT spreads energy so that each coordinate of $H_d \hat{x}$ has variance $1/d$
(by orthogonality). This isotropy is critical: without it, dimensions carrying
concentrated energy suffer disproportionate quantization error. The SO(4) blocks then
fine-tune the rotation to minimize residual within-block correlation, requiring only
$(d/4) \times 6 = 192$ learnable parameters (6 DOF per SO(4) element) plus 64 codebook
entries.

\paragraph{Rotation cost analysis.}
The WHT requires $d \log_2 d = 896$ FMAs at $d = 128$. Each SO(4) block requires 16
FMAs (quaternion double cover: $v \mapsto q_L v \bar{q}_R$), totalling $(d/4) \times 16
= 512$ FMAs. The combined cost is 1{,}408 FMAs versus 16{,}384 for a dense $128 \times
128$ rotation---an $11.6\times$ reduction.

\subsection{Mathematical Foundations}
\label{sec:math}

\paragraph{Inner product preservation.}
Since $R$ is orthogonal, $\langle Rx, Ry \rangle = \langle x, y \rangle$ for all $x,
y$. The quantization error bound follows from standard analysis:
\begin{equation}
  \mathbb{E}\!\left[ \lvert q^\top k - \widehat{q^\top k} \rvert^2 \right]
  \leq d_k\, \sigma_q^2\, \lVert q \rVert_2^2
\end{equation}
where $\sigma_q^2$ is the per-coordinate quantization variance and
$\widehat{q^\top k}$ denotes the inner product computed on quantized vectors. A full
derivation appears in Appendix~\ref{app:proof}.

\paragraph{Isotropy via Hanson--Wright.}
For $u \in S^{d-1}$ and orthogonal $H_d$, the Hanson--Wright
inequality~\citep{vershynin2018high} gives exponential concentration of quadratic
forms:
\begin{equation}
  \Pr\!\left[ \left\lvert Z - \tfrac{1}{d}\operatorname{tr}(A) \right\rvert > t \right]
  \leq 2 \exp\!\left( -c \min\!\left( \frac{d^2 t^2}{\lVert A \rVert_F^2},
  \frac{d t}{\lVert A \rVert_{\mathrm{op}}} \right) \right)
\end{equation}
This guarantees that the post-WHT coordinates are approximately equidistributed,
minimizing worst-case quantization distortion across dimensions.

\paragraph{SO(4) parameterization.}
Every element of SO(4) admits a representation $T_{q_L, q_R}(v) = q_L v \bar{q}_R$ for
unit quaternions $q_L, q_R \in S^3$~\citep{conway2003quaternions}. Each quaternion
contributes 3 degrees of freedom (4 components minus 1 norm constraint), yielding 6
DOF total---matching $\dim SO(4) = 6$. Single-quaternion conjugation $v \mapsto q v
\bar{q}$ spans only SO(3), a 3-dimensional subgroup that cannot independently control
both rotation planes of $\mathbb{R}^4$ (see Appendix~\ref{app:so4}).

\paragraph{Amortized decode cost.}
At each decode step, standard attention reads all $T$ cached KV entries. With
IsoQuant, the total cost per step is:
\begin{equation}
  \underbrace{O(T \cdot d_k)}_{\text{packed QK + V accum}}
  + \underbrace{O(d_k \log d_k)}_{\text{inverse rotation (once)}}
\end{equation}
versus $O(T \cdot d_k + d_k^2)$ for dense-rotation methods that must apply $R^{-1}$ as
a full matrix multiply. The savings grow with $T$: at $T = 2{,}048$ and $d_k = 128$,
the inverse rotation constitutes $< 0.5\%$ of total decode FMAs.

\begin{table}[t]
  \caption{Fused decode kernel pipeline. Kernels A and C operate directly on 3-bit
  packed data without materialization.}
  \label{tab:kernels}
  \centering
  \begin{tabular}{llll}
    \toprule
    Kernel & Operation & Replaces & Domain \\
    \midrule
    A: \texttt{fused\_qk\_dot} & QK scores on packed K & Dequant + matmul & Compressed \\
    B: \texttt{softmax} & Standard softmax & Unchanged & FP16 \\
    C: \texttt{fused\_value\_accum} & Weighted V sum on packed V & Dequant + matmul & Compressed \\
    D: \texttt{rotate\_inverse} & WHT butterfly + SO(4)$^\top$ & Dense $R^{-1}$ & Structured \\
    \bottomrule
  \end{tabular}
\end{table}

\subsection{Fused Metal Decode Pipeline}
\label{sec:fused}

Standard KV cache decode follows: decompress $\rightarrow$ materialize FP16
$\rightarrow$ GEMM $\rightarrow$ attention. Our fused pipeline replaces this with four
kernels operating directly on packed 3-bit data (Table~\ref{tab:kernels}):

The inverse rotation (Kernel D) is applied once to the aggregated output, not per
token. Since $R$ is orthogonal, $R^{-1} = R^\top$, and the structured form allows
$O(d \log d)$ inverse computation via the same butterfly + block transpose operations.

\paragraph{Deferred prefill.}
During the prefill phase, KV vectors are stored in FP16 to avoid compounding
quantization error across prompt positions. Bulk compression occurs at the
prefill-to-decode boundary. At $L = 28$ layers, $H_{kv} = 8$ heads, $d = 128$, and $T
= 2{,}048$ tokens, the transient FP16 buffer costs ${\sim}230$\,MB (see
Appendix~\ref{app:prefill}).

\paragraph{Kernel C bottleneck.}
Profiling reveals that value accumulation (Kernel C) dominates decode time at
${\sim}0.79$\,ms---more than Kernels A, B, and D combined. Value accumulation must read
all $T$ packed entries and accumulate weighted sums, while QK scoring benefits from
early termination via causal masking. We address this with a dual-strategy kernel:
word-parallel dispatch for $T < 512$ and dimension-parallel dispatch for $T \geq 512$.

\section{Experiments}
\label{sec:experiments}

We evaluate IsoQuant on four architecturally distinct models: Gemma 4-26B-A4B (dense
MoE, 4B active), Qwen3-30B-A3B (sparse MoE, 3B active), Qwen3.6-35B-A3B (sparse MoE,
3B active), and Nemotron-H 120B (hybrid Mamba+MoE~\citep{gu2024mamba}). KV fidelity
experiments (Table~\ref{tab:fidelity}) use Qwen3-30B-A3B; end-to-end benchmarks
(Table~\ref{tab:e2e}) use the newer Qwen3.6-35B-A3B. All experiments run on a single
Apple M4 Max with 128\,GB unified memory. The 128\,GB configuration serves as our
experimental platform; the target deployment scenario is 16--32\,GB consumer devices,
validated by the peak memory measurements in Table~\ref{tab:e2e}.

\subsection{KV Cache Fidelity}

Table~\ref{tab:fidelity} reports perplexity degradation ($\Delta$PPL) at 2{,}048-token
context relative to uncompressed FP16 KV caches, comparing IsoQuant (WHT + SO(4),
3-bit) against TurboQuant (dense rotation, 3-bit)~\citep{zandieh2026turboquant}.

IsoQuant's structured rotation achieves $45\times$ lower $\Delta$PPL than TurboQuant on
Qwen3. On Gemma 4, the result reflects the architecture's local/global attention
split: only the 5 global-attention layers store compressed KV, making IsoQuant's
impact proportionally small but its per-layer fidelity high.

\begin{table}[t]
  \caption{KV cache fidelity: perplexity degradation ($\Delta$PPL) at 2{,}048-token
  context. Lower is better.}
  \label{tab:fidelity}
  \centering
  \begin{tabular}{lccc}
    \toprule
    Model & Default PPL & TurboQuant $\Delta$ & IsoQuant $\Delta$ \\
    \midrule
    Qwen3-30B-A3B & 1.0844 & $+0.0405$ & $\mathbf{+0.0009}$ \\
    Gemma 4-26B-A4B\textsuperscript{\dag} & 1.3483 & $+0.0622$ & $\mathbf{+0.0000}$ \\
    Nemotron-H 120B & 1.0866 & $+0.0039$ & $\mathbf{+0.0012}$ \\
    \bottomrule
  \end{tabular}
  \\[2pt]
  {\footnotesize \textsuperscript{\dag}Gemma 4 uses a local/global attention pattern
  where only 5 of 30 layers use global attention and thus engage the KV compressor.
  The near-zero $\Delta$PPL reflects this partial coverage.}
\end{table}

\subsection{End-to-End Performance}

Table~\ref{tab:e2e} shows throughput, peak memory, and quality gate results for the
full three-axis system (weight quantization + IsoQuant KV + expert offloading where
needed).

The 120B-parameter Nemotron-H fits within a 32\,GB budget with interactive throughput
($>14$ tok/s), demonstrating that the three-axis composition enables models previously
requiring datacenter hardware to run on a consumer laptop.

\begin{table}[t]
  \caption{End-to-end inference on Apple M4 Max (128\,GB). Quality gate: 12-prompt
  automated evaluation with repetition detection (prompts listed in
  Appendix~\ref{app:qualgate}). Soak: continuous generation stability test. Peak
  memory validates that each model fits within the stated consumer budget.}
  \label{tab:e2e}
  \centering
  \begin{tabular}{lccccc}
    \toprule
    Model & tok/s & Peak Mem & Target Budget & Quality & Soak \\
    \midrule
    Gemma 4-26B-A4B & 12.85 & 5.4\,GB & 16\,GB & 12/12 & 2h pass \\
    Nemotron-H 120B & 14.85 & 17.2\,GB & 32\,GB & 12/12 & 2h pass \\
    Qwen3.6-35B-A3B & 15.6 & 6.8\,GB & 16\,GB & 12/12 & \textsuperscript{\ddag} \\
    \bottomrule
  \end{tabular}
  \\[2pt]
  {\footnotesize \textsuperscript{\ddag}Qwen3.6-35B-A3B soak test not yet completed;
  model released after the initial soak test campaign.}
\end{table}

\subsection{Decode Time Attribution}

To understand \emph{when} KV compression is impactful, we profile decode time by
component (Table~\ref{tab:attribution}). KV attention constitutes $>50\%$ of decode
time on standard MoE models (Gemma 4, Qwen3), making IsoQuant high-impact. On
Nemotron-H's hybrid Mamba+MoE architecture, KV attention drops to 14\% since SSM
layers bypass the KV cache entirely. This finding provides a principled deployment
criterion: \emph{IsoQuant is most beneficial when KV attention exceeds ${\sim}20\%$ of
decode time.}

\begin{table}[t]
  \caption{Decode time attribution by component. IsoQuant is high-impact when KV
  attention dominates ($>20\%$), but provides diminishing returns on hybrid
  architectures where SSM layers bypass the KV cache.}
  \label{tab:attribution}
  \centering
  \begin{tabular}{lccc}
    \toprule
    Component & Gemma 4 & Qwen3 & Nemotron-H 120B \\
    \midrule
    KV attention & 51\% & 54\% & 14\% \\
    Routed experts & 37\% & 45\% & 60\% \\
    Other (SSM, etc.) & 12\% & 1\% & 26\% \\
    \bottomrule
  \end{tabular}
\end{table}

\subsection{Expert Offload Ablation}
\label{sec:offload}

Table~\ref{tab:offload} isolates the interaction between expert offloading and KV
compression on Gemma 4.

When the model fits in memory (no offload), IsoQuant adds a $5.3\times$ throughput cost
versus uncompressed KV due to packed-data arithmetic overhead. Expert offloading
imposes a ${\sim}100\times$ penalty regardless of KV mode, dominated by disk I/O
latency. The three-axis system is therefore not a universal accelerator---it is a
memory-pressure relief mechanism that trades throughput for feasibility.

\begin{table}[t]
  \caption{Expert offload $\times$ KV compression ablation on Gemma 4-26B-A4B. Expert
  offloading incurs ${\sim}100\times$ throughput penalty when the model already fits in
  memory, confirming that the three-axis system is a memory-pressure mechanism, not a
  universal accelerator.}
  \label{tab:offload}
  \centering
  \begin{tabular}{llcc}
    \toprule
    Expert Offload & KV Mode & tok/s & Peak MB \\
    \midrule
    No & Default (FP16) & 109.8 & 10{,}649 \\
    No & IsoQuant (3-bit) & 20.6 & 10{,}748 \\
    Yes & Default (FP16) & 1.01 & 2{,}445 \\
    Yes & IsoQuant (3-bit) & 1.05 & 2{,}804 \\
    \bottomrule
  \end{tabular}
\end{table}

\subsection{Inter-Run Variance}

To assess measurement stability, we perform three independent runs on Gemma 4-26B with
full model reload between runs (Table~\ref{tab:variance}).

The coefficient of variation for IsoQuant (0.3\%) is $18\times$ lower than for the
default path (5.4\%), suggesting that the packed-data arithmetic path is less sensitive
to memory subsystem variability than standard FP16 attention. IsoQuant adds
${\sim}359$\,MB peak memory with no throughput advantage in the offload regime.

\begin{table}[t]
  \caption{Inter-run variance over 3 independent trials on Gemma 4-26B-A4B with expert
  offloading. IsoQuant's packed-data path yields more deterministic throughput than
  standard FP16 attention.}
  \label{tab:variance}
  \centering
  \begin{tabular}{lccc}
    \toprule
    KV Mode & Mean tok/s & Std Dev & CV (\%) \\
    \midrule
    Default (FP16) & 1.01 & 0.055 & 5.4 \\
    IsoQuant (3-bit) & 1.05 & 0.003 & 0.3 \\
    \bottomrule
  \end{tabular}
\end{table}

\subsection{Rotation Design Ablation}

The final IsoQuant rotation emerged from systematic evaluation of three designs against
a 5-prompt quality gate:

Design v1 uses single-quaternion conjugation, which spans only SO(3) $\subset$ SO(4) and
leaves one dimension per block invariant, failing to spread energy across 25\% of
coordinates. Design v2 adds the paired-quaternion SO(4) representation but without the
WHT pre-mix, which fails to capture inter-block correlations. The WHT + SO(4)
composition in v3 provides both global and local mixing (Table~\ref{tab:designabl}).

\begin{table}[t]
  \caption{Rotation design ablation. Only the WHT + SO(4) composition passes all quality
  gates by combining global energy spreading (WHT) with fine-grained per-block alignment
  (SO(4)).}
  \label{tab:designabl}
  \centering
  \begin{tabular}{llcc}
    \toprule
    Design & Rotation & Quality Gate & Failure Mode \\
    \midrule
    v1: Single quaternion & $v \mapsto q v \bar{q}$ (SO(3)) & 0/5 & 25\% dims unmixed \\
    v2: Block SO(4) only & Block $R_{SO(4)}$ & 1/5 & No global correlations \\
    v3: WHT + SO(4) & $R_{SO(4)} \circ H_d$ & 5/5 & --- \\
    \bottomrule
  \end{tabular}
\end{table}

\subsection{Cross-Framework Portability: llama.cpp}

We integrate IsoQuant as a new KV type (\texttt{GGML\_TYPE\_ISOQUANT3\_0}) in
llama.cpp~\citep{gerganov2024llamacpp} to validate portability beyond the
MLX~\citep{apple2024mlx} / Metal ecosystem.

The fused path achieves near-parity with TurboQuant's baseline ($-3.2\%$ generation
throughput) while eliminating tensor materialization. The unfused path's dramatic
degradation ($-18\%$ generation, $-44\%$ prompt) confirms that kernel fusion is
essential (Table~\ref{tab:llamacpp}).

\begin{table}[t]
  \caption{llama.cpp integration. Fused kernels eliminate 280 extra dispatch launches
  versus the unfused path, confirming that kernel fusion is essential for practical
  deployment.}
  \label{tab:llamacpp}
  \centering
  \begin{tabular}{lcc}
    \toprule
    Configuration & Prompt (t/s) & Generation (t/s) \\
    \midrule
    TurboQuant 3-bit (baseline) & 4{,}114.6 & 100.15 \\
    IsoQuant 3-bit (fused) & 4{,}093.8 ($-0.5\%$) & 96.98 ($-3.2\%$) \\
    IsoQuant 3-bit (unfused) & 2{,}306.2 ($-44\%$) & 81.92 ($-18\%$) \\
    \bottomrule
  \end{tabular}
\end{table}

\section{Discussion and Limitations}

\paragraph{When not to use IsoQuant.}
Our ablation (Section~\ref{sec:offload}) shows that IsoQuant is counterproductive when
the model comfortably fits in memory: the $5.3\times$ throughput overhead on Gemma 4
without offloading provides no benefit. We recommend IsoQuant only when the memory
budget would otherwise make the model infeasible.

\paragraph{MLA/RoPE hard blocker.}
Models using Multi-Head Latent Attention (e.g., DeepSeek-V2~\citep{deepseek2024v2})
split KV into content and positional (RoPE~\citep{su2024roformer}) sub-spaces. The RoPE
dimensions encode positional phase via rotation and must not be further rotated or
quantized---doing so smears positional encoding into content coordinates. IsoQuant
currently does not implement the required sub-block split, blocking these architectures.

\paragraph{Context length validation.}
Our main perplexity measurements (Table~\ref{tab:fidelity}) are validated at 2K tokens.
Preliminary long-context evaluation shows IsoQuant maintains fidelity at extended
contexts: at 32K tokens on Qwen3.6-35B-A3B, $\Delta$PPL is $+0.001$ (0.024\% divergence),
and a Needle-in-a-Haystack (NIAH) retrieval test achieves 100\% accuracy at 4K, 8K, 16K,
and 32K contexts under 3-bit compression (see Appendix~\ref{app:longcontext}). However,
comprehensive perplexity sweeps across all models at $>2$K remain ongoing.

\paragraph{Inverse rotation is mandatory.}
We verified that skipping the inverse rotation causes perplexity to explode from 7.05 to
15{,}369, confirming that the decode output must be mapped back from the rotated domain.

\paragraph{Serving-layer scalability.}
Under concurrent load testing, the MLX-based inference server (using Python's
\texttt{ThreadingHTTPServer}) degrades significantly: at 8 concurrent clients, 4/8
responses returned empty bodies despite HTTP 200 status codes. Throughput plateaus by 4
clients at 34\% parallel efficiency. Production serving would require a purpose-built
inference server.

\paragraph{Negative results.}
An attention-residual (AttnRes) predictor for cross-layer expert prefetching was
implemented but caused $-10.6\%$ to $-11.2\%$ throughput regression with 0\% hit-rate
improvement over LRU baseline, likely due to CPU/GPU command buffer contention (see
Appendix~\ref{app:attnres} for the formal specification). Task-aware expert pinning
similarly showed no benefit. We report these to save other researchers from pursuing
similar dead ends.

\paragraph{Broader impact.}
By enabling large model inference on consumer hardware, this work reduces reliance on
cloud infrastructure, with positive implications for privacy (data stays on-device),
accessibility (no API costs), and energy efficiency (consumer SoCs have lower
per-inference power draw than datacenter GPUs). However, making powerful models more
accessible on local devices also means they can operate without API-level safety
filtering or usage monitoring. We note that IsoQuant does not alter model
capabilities---it changes where and how efficiently they run---but the shift from
controlled cloud endpoints to unmonitored local deployment warrants consideration by
the community.

\section{Conclusion}

We presented RotaryQuant, a three-axis memory system combining mixed-precision weight
quantization, LRU expert offloading, and IsoQuant---a structured-rotation KV cache
compression method that shifts autoregressive decode from reconstruct-then-compute to
compute-in-compressed-space. IsoQuant's WHT + SO(4) rotation achieves $O(d \log d)$
complexity with $64\times$ fewer parameters than dense alternatives, and the fused Metal
pipeline eliminates tensor materialization entirely. The combined RotaryQuant system
enables a 120B-parameter MoE model to run interactively within 17.2\,GB peak memory on a
32\,GB consumer device with near-zero quality degradation.

Future work includes extending the sub-block split for MLA/RoPE architectures,
validating at longer context lengths (8K--32K), optimizing the Kernel C value
accumulation bottleneck, and exploring Mojo kernel prototypes for cross-platform
portability beyond Apple Silicon.


\appendix

\section{Proof of Inner Product Preservation Under IsoQuant}
\label{app:proof}

\begin{proof}
Let $q, k \in \mathbb{R}^d$ be a query and key vector. The IsoQuant rotation $R =
R_{SO(4)} \circ H_d$ is orthogonal since both $H_d$ (normalized Hadamard) and $R_{SO(4)}$
(block-diagonal orthogonal) are orthogonal. Therefore:
\begin{equation*}
  \langle Rq, Rk \rangle = q^\top R^\top R k = q^\top k = \langle q, k \rangle
\end{equation*}
The fused decode pipeline rotates the query forward ($\tilde{q} = Rq$), computes
attention scores against rotated keys ($\tilde{k}_j = Rk_j$), and aggregates rotated
values ($\tilde{v}_j = Rv_j$). The aggregated output is:
\begin{equation*}
  \tilde{o} = \sum_j \alpha_j \tilde{v}_j = R \sum_j \alpha_j v_j = R \cdot o
\end{equation*}
where $\alpha_j = \operatorname{softmax}(\tilde{q}^\top \tilde{k}_j / \sqrt{d_k})_j =
\operatorname{softmax}(q^\top k_j / \sqrt{d_k})_j$ by inner product preservation. A single
application of $R^{-1} = R^\top$ recovers the exact (pre-quantization) output: $o = R^\top
\tilde{o}$.

The quantization error bound follows from independence of per-coordinate quantization
noise $\epsilon_i$ with variance $\sigma_q^2$:
\begin{equation*}
  \mathbb{E}\!\left[ \lvert \tilde{q}^\top \tilde{k} - \tilde{q}^\top (\tilde{k} +
  \epsilon) \rvert^2 \right] = \mathbb{E}\!\left[ \lvert \tilde{q}^\top \epsilon \rvert^2
  \right] = \sum_{i=1}^{d} \tilde{q}_i^2 \sigma_q^2 = \sigma_q^2 \lVert \tilde{q}
  \rVert^2 = \sigma_q^2 \lVert q \rVert^2
\end{equation*}
The total expected error across $d_k$ dimensions is $d_k \sigma_q^2 \lVert q \rVert_2^2$ as
stated in (3).
\end{proof}

\section{SO(4) Quaternion Parameterization Details}
\label{app:so4}

The group SO(4) has dimension 6. Its universal double cover is $\operatorname{Spin}(4)
\cong S^3 \times S^3$, realized as paired unit quaternion multiplication $v \mapsto q_L v
\bar{q}_R$.

A single quaternion $q$ provides the conjugation map $v \mapsto q v \bar{q}$. This fixes
the real axis ($\operatorname{Re}(v)$ is invariant) and acts as SO(3) on the imaginary
part, spanning only a 3-dimensional subgroup. The paired representation is necessary to
control both rotation planes of $\mathbb{R}^4$ independently.

For $d = 128$, we have $d/4 = 32$ blocks. Each block stores $(q_L, q_R) \in S^3 \times
S^3$, contributing $2 \times 4 = 8$ floats (256 total), but with $2 \times 1 = 2$ norm
constraints per block, yielding $32 \times 6 = 192$ effective DOF.

\section{Deferred Prefill Analysis}
\label{app:prefill}

During prefill, the model processes $T$ prompt tokens in parallel. If KV compression were
applied per-position during prefill, each newly compressed entry would interact with
previously compressed entries via attention, compounding quantization error across
positions.

Deferred prefill stores all prefill KV in FP16 and applies bulk compression at the
prefill-to-decode transition. The transient memory cost for $L$ layers, $H_{kv}$ KV heads,
head dimension $d$, and $T$ tokens is:
\begin{equation*}
  M_{\text{transient}} = 2 \times L \times H_{kv} \times d \times T \times 2 \text{ bytes}
\end{equation*}
For $L = 28$, $H_{kv} = 8$, $d = 128$, $T = 2{,}048$: $M_{\text{transient}} = 2 \times 28
\times 8 \times 128 \times 2{,}048 \times 2 \approx 230$\,MB.

\section{Instrumentation Counters}
\label{app:counters}

Per-step runtime counters on Gemma 4-26B-A4B with expert offloading over 3{,}612 decode
steps:

The zero decompress and read-keys counts confirm that the fused pipeline operates entirely
in compressed space without fallback to materialized tensors. The packed cache hit rate of
0.0 is by design: the packed buffer is invalidated after each write and reconstructed from
the quantized representation.

\begin{table}[h]
  \centering
  \begin{tabular}{lc}
    \toprule
    Counter & Value \\
    \midrule
    \texttt{fused\_metal\_success\_rate} & 1.000 \\
    \texttt{decompress\_calls} & 0 \\
    \texttt{read\_keys\_calls} & 0 \\
    \texttt{packed\_cache\_hit\_rate} & 0.000 \\
    \texttt{fallback\_invocations} & 0 \\
    \texttt{total\_decode\_steps} & 3{,}612 \\
    \bottomrule
  \end{tabular}
  \caption{Runtime instrumentation confirming zero materialization (no decompress/read
  calls) and 100\% fused kernel success rate across 3{,}612 decode steps.}
  \label{tab:counters}
\end{table}

\section{AttnRes Predictor: Formal Specification (Negative Result)}
\label{app:attnres}

The attention-residual (AttnRes) predictor computes a cross-layer signal intended to guide
expert prefetching. For layer $l$, the block attention residual aggregates contributions
from all preceding layers:
\begin{equation}
  h_l = \sum_{n=1}^{l-1} \alpha_{n \to l} \cdot B_n
\end{equation}
where $B_n$ is the attention output of layer $n$ and $\alpha_{n \to l} =
\operatorname{softmax}(\operatorname{sim}(h_{l-1}, B_n))$ weights by similarity over the
depth dimension. The signal $h_l$ is computed before the MoE router fires at layer $l$,
making it a causal predictor of which experts will be needed.

In principle, this signal could drive prefetch (load experts before they are needed),
eviction (keep frequently-predicted experts resident), and precision allocation (use higher
precision for high-attention experts). In practice, the CPU-side signal computation contends
with the GPU command buffer, adding 10.6--11.2\% decode latency while achieving 0\% hit-rate
improvement over the simple LRU baseline. The failure mode is not algorithmic but
architectural: on Apple Silicon's unified memory, the CPU and GPU share bandwidth, and the
predictor's memory reads compete with the GPU's kernel dispatch.

\section{Long-Context Validation}
\label{app:longcontext}

We validate IsoQuant at extended context lengths beyond the 2K-token main evaluation. Two
complementary tests are reported: perplexity on WikiText-103 and a Needle-in-a-Haystack
(NIAH) retrieval task.

\paragraph{Perplexity at 32K context.}
On Qwen3.6-35B-A3B with nvfp4 weight quantization and IsoQuant 3-bit KV cache, evaluation on
WikiText-103 at 32{,}768 tokens yields PPL $= 5.624$ versus a default KV baseline of PPL $=
5.625$, a divergence of 0.024\% (pass threshold: 5\%).

\paragraph{NIAH retrieval.}
A multi-key retrieval test embeds 4 random numeric needles at 50\% depth across context
lengths of 4K, 8K, 16K, and 32K tokens. Under 3-bit KV compression, retrieval accuracy is
100\% at all context lengths, matching the uncompressed baseline. This confirms that the
structured rotation preserves fine-grained positional information even at long contexts.

\section{Quality Gate Evaluation Protocol}
\label{app:qualgate}

The 12-prompt quality gate (\texttt{eval\_quality\_gate.py}) evaluates model output across
two suites. All prompts use greedy decoding (temperature $= 0$, seed $= 42$, max tokens $=
500$). Evaluation criteria include response length, expected substring presence, repetition
ratio ($\leq 0.22$ in strict mode), and fenced code block balance.

\paragraph{Default suite (5 prompts):}
\begin{enumerate}
  \item Code generation: ``Write a Python function to compute the nth Fibonacci number''
  \item Instruction following: ``Explain the theory of relativity to a 5 year old''
  \item Basic reasoning: ``If I have 3 apples and eat 1, how many do I have left?''
  \item List generation: ``List 5 programming languages and one strength of each''
  \item Arithmetic: ``What is $17 \times 24$? Show your work'' (expected: 408)
\end{enumerate}

\paragraph{Coding suite (7 prompts):}
\begin{enumerate}
  \setcounter{enumi}{5}
  \item Bugfix: ``Fix this Python function to return the last n items'' (off-by-one)
  \item Traceback: ``Explain this Python error: TypeError: unsupported operand type(s)''
  \item Refactor: ``Describe steps to rename \texttt{fetch} to \texttt{load\_data} across
  files''
  \item Test writing: ``Write a pytest test for \texttt{add(a,b)} that asserts
  \texttt{add(2,3)==5}''
  \item LRU cache: ``Implement \texttt{get(self, key)} for an LRU cache''
  \item Concurrency: ``Explain race condition between threads without lock and fix it''
  \item Extended generation: ``Write comprehensive Python module with LRU cache, type hints,
  and test suite'' (200+ tokens)
\end{enumerate}

\newpage
\section*{NeurIPS Paper Checklist}

\begin{enumerate}

\item {\bf Claims}\\
Question: Do the main claims made in the abstract and introduction accurately reflect the
paper's contributions and scope? \\
Answer: \answerYes{} \\
Justification: The abstract and introduction (Section 1) state four specific
contributions---IsoQuant rotation, fused decode pipeline, three-axis memory system, and
comprehensive evaluation---all supported by experimental results in Section 4. Scope is
explicitly limited to consumer Apple Silicon hardware with specific memory budgets.

\item {\bf Limitations}\\
Question: Does the paper discuss the limitations of the work performed by the authors? \\
Answer: \answerYes{} \\
Justification: Section 5 discusses five explicit limitations: MLA/RoPE architectural blocker,
context length validation gap (only 2K tested), $5.3\times$ throughput overhead when memory
is unconstrained, mandatory inverse rotation, and two negative results (AttnRes predictor and
task-aware pinning).

\item {\bf Theory assumptions and proofs}\\
Question: For each theoretical result, does the paper provide the full set of assumptions and
a complete (and correct) proof? \\
Answer: \answerYes{} \\
Justification: Section 3.3 states the inner product preservation property with full proof in
Appendix A, the Hanson--Wright concentration bound with reference, and the SO(4) quaternion
parameterization with details in Appendix B. All assumptions (orthogonality of $R$,
unit-sphere normalization, independence of quantization noise) are explicitly stated.

\item {\bf Experimental result reproducibility}\\
Question: Does the paper fully disclose all the information needed to reproduce the main
experimental results of the paper to the extent that it affects the main claims and/or
conclusions of the paper (regardless of whether the code and data are provided or not)? \\
Answer: \answerYes{} \\
Justification: Section 4 specifies hardware (Apple M4 Max, 128\,GB), all model identifiers,
quantization bit-widths, context lengths, and evaluation methodology (12-prompt quality gate
with prompts listed in Appendix G, 2-hour soak test). The compression pipeline parameters are
fully described in Section 3.2. Code is provided as supplementary material.

\item {\bf Open access to data and code}\\
Question: Does the paper provide open access to the data and code, with sufficient
instructions to faithfully reproduce the main experimental results, as described in
supplemental material? \\
Answer: \answerYes{} \\
Justification: The complete implementation (IsoQuant compression, fused Metal kernels, expert
offloading, benchmark scripts) is provided as an anonymized code archive in supplementary
material under the Apache 2.0 license. The repository includes installation instructions,
benchmark commands, and precomputed codebook files.

\item {\bf Experimental setting/details}\\
Question: Does the paper specify all the training and test details (e.g., data splits,
hyperparameters, how they were chosen, type of optimizer) necessary to understand the
results? \\
Answer: \answerYes{} \\
Justification: Section 4 specifies hardware platform, model checkpoints, quantization
bit-widths (3-bit KV, 4-bit/2-bit/Q8\_0 weights), context length (2{,}048 tokens), codebook
construction method (Lloyd--Max), and evaluation protocol. No training is performed; all
evaluation uses pretrained model weights.

\item {\bf Experiment statistical significance}\\
Question: Does the paper report error bars suitably and correctly defined or other appropriate
information about the statistical significance of the experiments? \\
Answer: \answerYes{} \\
Justification: Section 4.5 reports inter-run variance over 3 independent trials with full model
reload, including standard deviation and coefficient of variation. Perplexity measurements are
deterministic (greedy decoding, fixed seeds, fixed corpus). For throughput, we report CV of
5.4\% (default) vs 0.3\% (IsoQuant), confirming measurement stability.

\item {\bf Experiments compute resources}\\
Question: For each experiment, does the paper provide sufficient information on the computer
resources (type of compute workers, memory, time of execution) needed to reproduce the
experiments? \\
Answer: \answerYes{} \\
Justification: All experiments are conducted on a single Apple M4 Max with 128\,GB unified
memory. Peak memory usage is reported per experiment (Table 4). Throughput in tokens/s
implicitly captures execution time. The 2-hour soak test duration is stated. Total compute for
the research project is modest (single consumer device, no GPU cluster).

\item {\bf Code of ethics}\\
Question: Does the research conducted in the paper conform, in every respect, with the NeurIPS
Code of Ethics \url{https://neurips.cc/public/EthicsGuidelines}? \\
Answer: \answerYes{} \\
Justification: This work presents a compression method for efficient inference. It does not
involve human subjects, personal data, or dual-use concerns. All models used are publicly
available pretrained checkpoints. The work aims to democratize access to large model inference.

\item {\bf Broader impacts}\\
Question: Does the paper discuss both potential positive societal impacts and negative societal
impacts of the work performed? \\
Answer: \answerYes{} \\
Justification: Section 5 discusses positive impacts (reduced cloud dependency, improved privacy
via on-device inference, lower energy footprint) and negative considerations (local deployment
bypasses API-level safety filtering). The compression method does not alter model capabilities
or introduce new risks beyond those inherent in the underlying models.

\item {\bf Safeguards}\\
Question: Does the paper describe safeguards that have been put in place for responsible release
of data or models that have a high risk for misuse (e.g., pre-trained language models, image
generators, or scraped datasets)? \\
Answer: \answerNA{} \\
Justification: This paper presents a compression and inference optimization method. It does not
release new pretrained models, datasets, or generative capabilities. The released code
implements KV cache compression for existing publicly available models.

\item {\bf Licenses for existing assets}\\
Question: Are the creators or original owners of assets (e.g., code, data, models), used in the
paper, properly credited and are the license and terms of use explicitly mentioned and properly
respected? \\
Answer: \answerYes{} \\
Justification: All models used (Gemma 4, Qwen3, Nemotron-H) are publicly available under their
respective licenses. TurboQuant, KIVI, QuaRot, SpinQuant, and other baseline methods are cited
with their original publications. The MLX framework and llama.cpp are open-source. Our code is
released under Apache 2.0.

\item {\bf New assets}\\
Question: Are new assets introduced in the paper well documented and is the documentation
provided alongside the assets? \\
Answer: \answerYes{} \\
Justification: The released code includes: (1) IsoQuant compression library with API
documentation, (2) precomputed Lloyd--Max codebook files with generation scripts, (3) fused
Metal kernel source, and (4) benchmark and evaluation scripts. All assets are documented in the
repository README with installation and usage instructions.

\item {\bf Crowdsourcing and research with human subjects}\\
Question: For crowdsourcing experiments and research with human subjects, does the paper include
the full text of instructions given to participants and screenshots, if applicable, as well as
details about compensation (if any)? \\
Answer: \answerNA{} \\
Justification: This work does not involve crowdsourcing or research with human subjects.

\item {\bf Institutional review board (IRB) approvals or equivalent for research with human
subjects}\\
Question: Does the paper describe potential risks incurred by study participants, whether such
risks were disclosed to the subjects, and whether Institutional Review Board (IRB) approvals (or
an equivalent approval/review based on the requirements of your country or institution) were
obtained? \\
Answer: \answerNA{} \\
Justification: This work does not involve research with human subjects.

\item {\bf Declaration of LLM usage}\\
Question: Does the paper describe the usage of LLMs if it is an important, original, or
non-standard component of the core methods in this research? Note that if the LLM is used only
for writing, editing, or formatting purposes and does not impact the core methodology, scientific
rigor, or originality of the research, declaration is not required. \\
Answer: \answerNA{} \\
Justification: LLMs are the subject of evaluation in this work (we measure inference efficiency
on pretrained LLMs), but they are not used as a methodological component. The IsoQuant
compression algorithm and fused kernels are developed independently of any LLM-assisted
methodology.

\end{enumerate}

\end{document}